\documentclass{article}

\usepackage{microtype}
\usepackage{graphicx}
\usepackage{subcaption}
\usepackage{booktabs} 
\usepackage{threeparttable}

\usepackage{hyperref}

\usepackage[accepted]{icml2025}

\usepackage{amsmath}
\usepackage{amssymb}
\usepackage{mathtools}
\usepackage{amsthm}

\usepackage[capitalize,noabbrev]{cleveref}

\theoremstyle{plain}

\theoremstyle{definition}

\theoremstyle{remark}

\usepackage[textsize=tiny]{todonotes}

\icmltitlerunning{Bridging Quantum and Classical Computing in Drug Design}

\begin{document}

\twocolumn[
\icmltitle{Bridging Quantum and Classical Computing in Drug Design: \\Architecture Principles for Improved Molecule Generation}



\icmlsetsymbol{equal}{*}

\begin{icmlauthorlist}
\icmlauthor{Andrew Smith}{yyy}
\icmlauthor{Erhan Guven}{yyy}

\end{icmlauthorlist}

\icmlaffiliation{yyy}{Whiting School of Engineering, Johns Hopkins University, Baltimore, Maryland}

\icmlcorrespondingauthor{Andrew Smith}{asmit397@jh.edu}

\icmlkeywords{Machine Learning, ICML, Drug Discovery}

\vskip 0.3in
]


\printAffiliationsAndNotice{}

\begin{abstract}
Hybrid quantum-classical machine learning offers a path to leverage noisy intermediate-scale quantum (NISQ) devices for drug discovery, but optimal model architectures remain unclear. We systematically optimize the quantum-classical bridge architecture of generative adversarial networks (GANs) for molecule discovery using multi-objective Bayesian optimization. Our optimized model (BO-QGAN) significantly improves performance, achieving a 2.27-fold higher Drug Candidate Score (DCS) than prior quantum-hybrid benchmarks and 2.21-fold higher than the classical baseline, while reducing parameter count by more than 60\%. Key findings favor layering multiple (3-4) shallow (4-8 qubit) quantum circuits sequentially, while classical architecture shows less sensitivity above a minimum capacity. This work provides the first empirically-grounded architectural guidelines for hybrid models, enabling more effective integration of current quantum computers into pharmaceutical research pipelines.
\end{abstract}

\section{Background and Significance}
Developing new pharmaceuticals is a notoriously long and expensive process, often exceeding a decade and costing billions of dollars \cite{hughes_principles_2011, singh_drug_2023}. Computational methods offer a path to accelerate this process, but identifying promising candidates within the vast chemical space (estimated at \(10^{60}\) molecules) remains a significant hurdle \cite{bassani_past_2023}.

Machine learning techniques, particularly graph neural networks like MolGAN \cite{de_cao_molgan_2022}, help navigate chemical space by learning from known molecules to generate novel candidates. However, challenges persist in effectively representing complex structures and generalizing from limited approved drug data (fewer than 3,000 small-molecule drugs have received FDA approval \cite{knox_drugbank_2024}), often leading to high bias or mode collapse.

Quantum computing offers a potential advantage by naturally modeling quantum mechanical effects (superposition, entanglement) crucial to molecular behavior \cite{cao_quantum_2019, mcardle_quantum_2020, feynman_simulating_1982}. Quantum neural networks (QNNs), operating on quantum states, could theoretically learn molecular physics more efficiently \cite{cao_quantum_2019}. However, current noisy intermediate-scale quantum (NISQ) hardware suffers from limited qubit counts, substantial noise, and restricted access, hindering practical application \cite{combarro_practical_2023, preskill_quantum_2018}.

\subsection{Hybrid Quantum-Classical Models: A Pragmatic Approach}
Hybrid quantum-classical models offer a pragmatic solution, combining smaller quantum circuits (capturing key quantum features) with larger classical networks (handling scale and decoding). Parameterized quantum circuits, which are analogous to classical layers, contain tunable parameters and have adjustable width (qubits) and depth (layers) \cite{benedetti_parameterized_2019}. A quantum computation result is bridged into the classical network, leveraging quantum expressivity within current hardware limits. We use Generative Adversarial Networks (GANs) as our framework due to established classical (MolGAN) and hybrid \cite{li_quantum_2021} benchmarks, allowing focused analysis of the hybrid architecture itself.

The original hybrid model for this task, QGAN-HG \cite{li_quantum_2021}, adapted MolGAN \cite{de_cao_molgan_2022} and matched its performance with significantly fewer parameters. Subsequent work explored variations like Wasserstein loss and gradient penalty integration \cite{jain_hybrid_2022}, different quantum component placements \cite{kao_exploring_2023}, and advanced GAN structures like CycleGAN \cite{anoshin_hybrid_2024}. The potential real-world applicability was highlighted by work generating and experimentally validating novel KRAS inhibitor candidates discovered using quantum-enhanced methods \cite{ghazi_vakili_quantum-computing-enhanced_2025}. While valuable, these studies often focus on framework modifications, overlooking architectural details. To isolate the impact of the quantum-classical interface architecture, we benchmark directly against the original MolGAN and QGAN-HG, anticipating that our derived principles may benefit more advanced frameworks later.

Designing the optimal quantum-classical architecture involves complex trade-offs between quantum expressivity, NISQ hardware limitations, and overall model parameter count. Establishing systematic principles to balance these factors for effective information transfer and performance is crucial, but currently lacks rigorous investigation.

This research systematically optimizes the quantum-classical hybrid architecture, hypothesizing that a principled approach can significantly enhance performance and efficiency. We aim to:
\begin{enumerate}
    \item Systematically evaluate the impact of quantum and classical component sizes on model performance using multi-objective Bayesian optimization.
    \item Derive design principles for quantum circuits within hybrid generators by analyzing optimized configurations.
\end{enumerate}
By establishing the first empirically grounded architectural guidelines, this work demonstrates substantial performance gains (2.3x $DCS$ increase over previous hybrid work) with reduced computational cost ($>$60\% fewer parameters than the classical baseline), offering a practical path to integrate NISQ computers more effectively into drug discovery.

\section{Materials and Methods}
\subsection{Molecular Representations as Graphs} \label{mol_graphs}
Molecules are represented as graphs \(\mathcal{G}(\mathcal{V}, \mathcal{E})\) with up to \(N=9\) heavy atoms (C, O, N, F) plus hydrogens from the QM9 dataset. Nodes \(v_i \in \mathcal{V}\) represent atoms with feature vectors \(x_i \in \mathbb{R}^{T=6}\) (one-hot encoding C, O, N, F, H, plus padding). Edges \((v_i, v_j) \in \mathcal{E}\) represent bonds with types \(y \in \{ \text{none, single, double, triple, aromatic} \}\) (\(Y=5\)). The graph is encoded by a feature matrix \(X \in \mathbb{R}^{N \times T}\) and an adjacency tensor \(A \in \mathbb{R}^{N \times N \times Y}\) indicating bond types \cite{de_cao_molgan_2022}. Molecules with fewer than \(N\) atoms are padded.

\subsection{Hybrid Generative Adversarial Network (GAN) Architecture}
\begin{figure}[htbp]
    \centering
    \includegraphics[width=0.8\columnwidth]{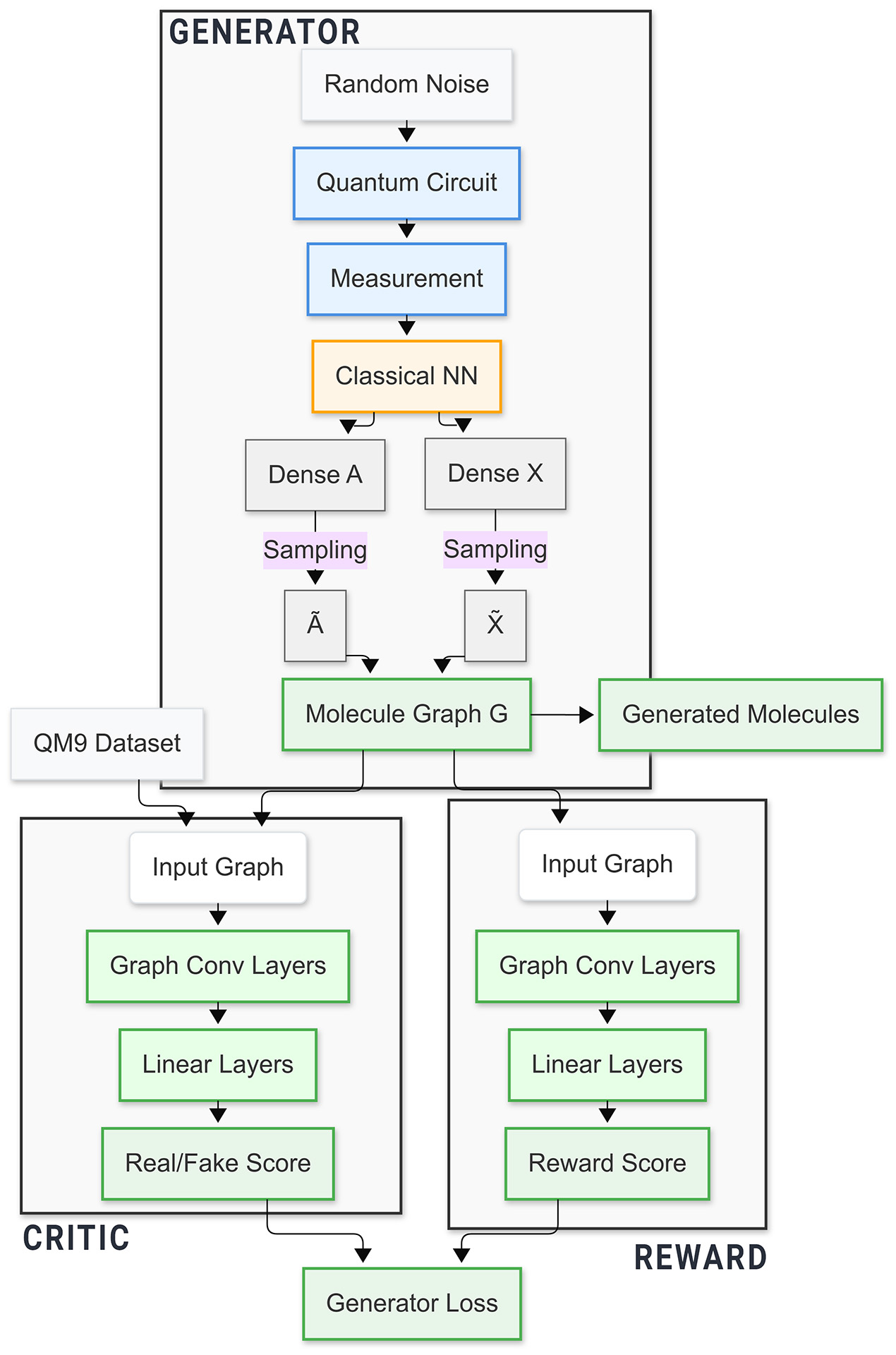}
    \caption{Architecture of the hybrid quantum-classical generator GAN with reward network.}
    \label{fig:arch}
\end{figure}

Our model adapts MolGAN \cite{de_cao_molgan_2022}, comprising a hybrid generator \(G_\theta\), a classical discriminator (critic), and a classical reward network (Figure \ref{fig:arch}).

\subsubsection{Generator Network (\(G_\theta\))}
The generator maps noise \(z \in \mathbb{R}^{B \times M}\) (batch size \(B\), quantum width \(M\)) to a molecular graph \(\mathcal{G}\). The input \(z\) is embedded into a quantum state using angle encoding. This state is transformed by parameterized quantum circuit layers consisting of single-qubit rotations (RY gates) and CNOT-based entangling layers (with a ring structure). Measurement of Pauli-Z expectation values yields a quantum latent representation (\(\mathbb{R}^{B \times M}\)). This representation is transferred to the classical component, where fully connected layers (with \(\tanh\) activations) output dense tensors \(A\) and \(X\). Gumbel-Softmax sampling produces discrete graph tensors \(\widetilde{A}\) and \(\widetilde{X}\). Key optimized hyperparameters are quantum width (\(M\)), quantum depth (number of circuit layers), classical width (neurons/layer), and classical depth (number of layers).

\subsubsection{Discriminator (Critic) Network}
The discriminator is a classical graph neural network, which distinguishes real vs. generated molecules (\(\widetilde{A}, \widetilde{X}\)), outputting a scalar Wasserstein critic score \(r\). It uses graph convolutions and aggregation layers.

\subsubsection{Reward Network}
Architecturally similar to the discriminator but with independent parameters, the reward network predicts the expected reward (desirable chemical properties, Section \ref{sec:metrics}) for a generated molecule \(\mathcal{G}\), guiding the generator via differentiable gradients \cite{de_cao_molgan_2022}.

\subsubsection{Training Details}
The generator loss is a weighted sum of the adversarial loss from the discriminator and the value loss from the reward network, controlled by a hyperparameter \(\lambda \in [0, 1]\). The Generator/Reward and Discriminator networks use separate Adam optimizers (LR \(1 \times 10^{-4}\)). Gradient clipping (norm 1.0) and a warm-up/constant/decay LR scheduler are used. Quantum circuits are implemented via PennyLane \cite{bergholm_pennylane_2022} with PyTorch \cite{paszke_pytorch_2019}.

\subsection{Evaluation Metrics} \label{sec:metrics}
Performance is assessed by realism and druglikeness, calculated using RDKit \cite{landrum_rdkitrdkit_2024}. Chemically invalid molecules score zero.
\begin{itemize}
    \item \textbf{Realism (Fréchet Distance, $FD$):} Measures distributional similarity between generated and real molecule batches using discrete Fréchet distance \cite{thomas_eiter_computing_1994, denaxas_spirosdiscrete_frechet_2019}. \(FD < 12.5\) indicates sufficient realism \cite{li_quantum_2021}.
    \item \textbf{Druglikeness ($QED$, $logP$, $SA$):}
        \begin{itemize}
            \item $QED$: Quantitative Estimation of Druglikeness, similarity to known oral drugs based on physicochemical properties \cite{bickerton_quantifying_2012}.
            \item $logP$: Octanol-water partition coefficient (lipophilicity) \cite{wildman_prediction_1999, lipinski_experimental_2001}.
            \item $SA$: Synthetic Accessibility score, estimating synthesis ease based on fragment contributions and complexity penalties \cite{ertl_estimation_2009}.
        \end{itemize}
    \item \textbf{Drug Candidate Score ($DCS$):} A composite metric designed to balance the above properties, calculated as:
    \begin{displaymath}
    DCS = 10 \times QED \times logP \times SA
    \end{displaymath}
    where 10 is a scaling factor for readability. Individual metrics ($QED, logP, SA$) are normalized to \([0, 1]\) (higher is better; note $SA$ normalization reverses standard interpretation).
\end{itemize}

\subsection{Bayesian Optimization of Bridge Architecture}

\begin{figure}[tb]
    \centering
    \includegraphics[width=0.7\columnwidth]{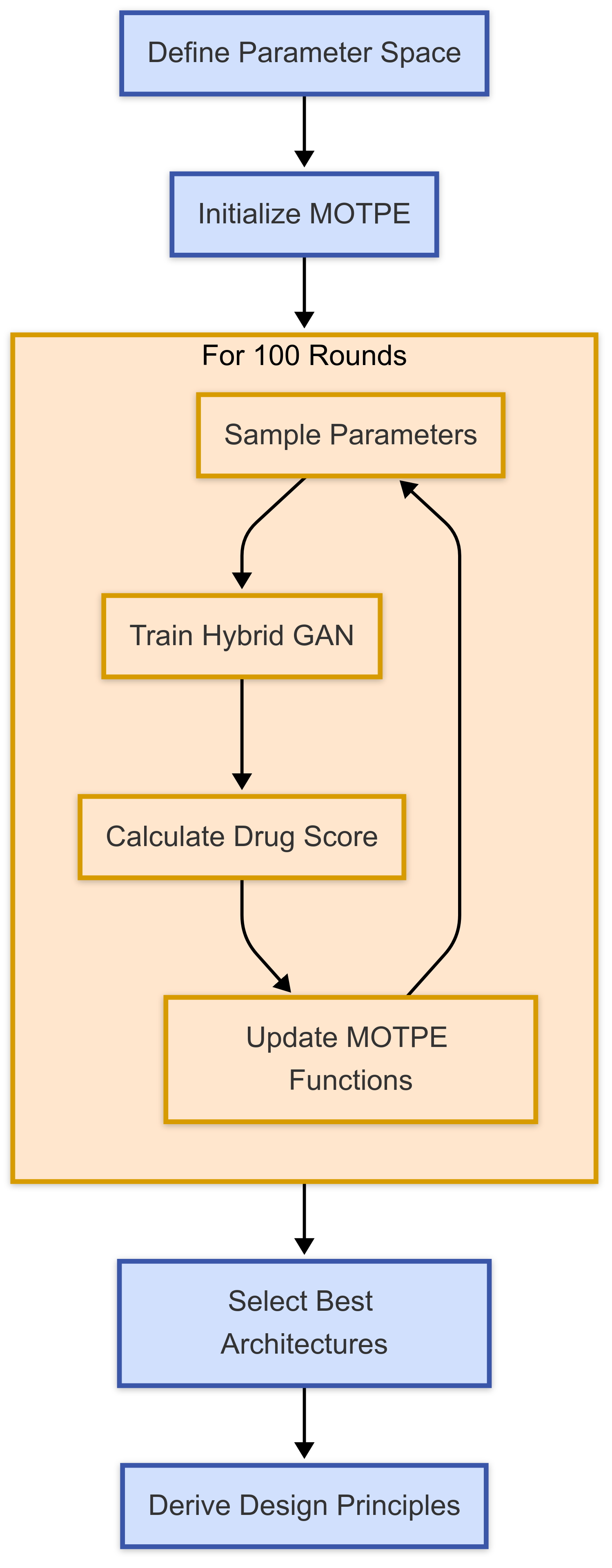}
    \caption{Workflow for optimizing the architecture with Multi-Objective Tree-structured Parzen Estimator (MOTPE).}
    \label{fig:optimizer}
\end{figure}

The optimization process used the Multi-Objective Tree-structured Parzen Estimator (MOTPE) algorithm, implemented via the Optuna library \cite{akiba_optuna_2019}. MOTPE is well suited for computationally expensive black-box optimization problems such as hyperparameter tuning \cite{ozaki_multiobjective_2020}. It builds probabilistic surrogate models (\(l(x)\) for promising parameters, \(g(x)\) for unpromising ones) based on Pareto dominance relationships from evaluated trials. It then suggests new hyperparameter configurations by maximizing the expected improvement criterion, efficiently exploring the search space with relatively few trials. Each optimization trial involved sampling an architecture, training the corresponding hybrid GAN model, and evaluating its performance ($DCS$ and training time), which is shown in Figure \ref{fig:optimizer}.

The search space covered quantum width (\([4, 16]\) qubits), quantum depth (\([1, 4]\) layers), classical width (\([16, 512]\) neurons), and classical depth (\([1, 4]\) layers).

\subsection{Dataset and Preprocessing}
We used the QM9 dataset (133,885 small organic molecules, up to 9 heavy atoms C, O, N, F) \cite{ramakrishnan_quantum_2014}, preprocessed into graph representations (Section \ref{mol_graphs}) including explicit hydrogens. Padding was used for molecules $<$ 9 heavy atoms. RDKit handled molecule processing \cite{landrum_rdkitrdkit_2024}.

\subsection{Experimental Design}
Phase 1 involved 100 MOTPE trials that optimized the architecture. Each trial trained a hybrid GAN, using a simulated quantum backend for up to 5000 epochs with early stopping (no validation $DCS$ improvement for 250 epochs after reaching $FD < 12.5$). Phase 2 benchmarked the best $DCS$ architecture discovered during Phase 1 against re-implemented MolGAN \cite{de_cao_molgan_2022} and QGAN-HG MR \cite{li_quantum_2021} using identical protocols. Performance metrics were averaged over 30 independent runs (1000 molecules/run). Statistical significance was assessed using t-tests, and effect size was assessed using Cohen's d. Experiments used Python 3.12, PyTorch 2.4.1, PyTorch Geometric, PennyLane, Optuna, RDKit \cite{paszke_pytorch_2019,fey_fast_2019, bergholm_pennylane_2022, akiba_optuna_2019, landrum_rdkitrdkit_2024}.

The results of a hybrid quantum-classical model discovered by the optimization process were then validated on real quantum hardware.

\section{Results}
\subsection{Bayesian Optimization Performance}
The MOTPE algorithm efficiently navigated the hyperparameter space in the 100 trials, progressively identifying improved architectures $DCS$. The algorithm effectively explored the trade-off between $DCS$ and training time, identifying a Pareto front of optimal configurations. The final model, which we call Bayesian-Optimized Quantum GAN (BO-QGAN), was selected from this front to maximize $DCS$.

\subsection{Quantitative Performance Comparison} \label{quant_perf}
Systematic optimization yielded substantial performance gains compared to baseline models. Table \ref{table:drug_properties} summarizes the key performance metrics for the proposed Bayesian Optimized QGAN (BO-QGAN) alongside both reported and recreated results for classical MolGAN \cite{de_cao_molgan_2022} and the baseline hybrid model, QGAN-HG MR \cite{li_quantum_2021}.

Averaged over 30 independent trials generating 1000 molecules each, our optimized architecture (BO-QGAN) achieved a mean $DCS$ of 1.190 $\pm$ 0.132. This substantially outperforms the recreated comparison models under identical evaluation conditions. Using independent two-sample t-tests, we found significant (p $<$ 0.001) improvements over both the recreated QGAN-HG MR (0.524 $\pm$ 0.083 $DCS$) and the recreated MolGAN (0.539 $\pm$ 0.068 $DCS$). This represents a 2.27-fold improvement over the baseline hybrid model and a 2.21-fold improvement over the classical baseline, with large effect sizes (Cohen's d $>$ 0.8) indicating a meaningful advance in generated candidate quality.

Beyond the composite $DCS$, the improvements in individual properties like lipophilicity ($logP$) and synthetic accessibility ($SA$) are particularly noteworthy (Table \ref{table:drug_properties}). The higher normalized $logP$ suggests molecules with properties more amenable to absorption and distribution according to Lipinski's guidelines \cite{lipinski_experimental_2001}, while the significantly improved $SA$ (0.33 vs. 0.21/0.23) indicates that the molecules generated by BO-QGAN are predicted to be considerably easier to synthesize in a lab, a critical factor for practical drug development \cite{ertl_estimation_2009}.

This enhanced performance was achieved with high efficiency: BO-QGAN uses only 158,252 total trainable parameters (158,231 classical and 21 quantum), achieving a greater than 60\% reduction compared to the classical MolGAN's 400,203 parameters.

\begin{table*}[htb]
    \centering
        \begin{threeparttable}
            \caption{Drug Properties from Inference of 1000 Molecules with various GAN Architectures.}
            \label{table:drug_properties}
            \begin{tabular}{lcccc}
                \toprule
                \textbf{Method} & \textbf{$QED$} & \textbf{$logP$} & \textbf{$SA$} & \textbf{$DCS$} \\
                \midrule
                MolGAN (reported) \cite{de_cao_molgan_2022} & \textbf{0.51}  & 0.66  & 0.08  & 0.27 \\
                MolGAN (recreated) & 0.39  & 0.60  & 0.23  & 0.54 \\
                QGAN-HG MR (reported) \cite{li_quantum_2021} & \textbf{0.51}  & 0.49  & 0.11  & 0.27 \\
                QGAN-HG MR (recreated) & 0.44  & 0.57  & 0.21  & 0.52 \\
                \midrule
                BO-QGAN (proposed) & 0.44  & \textbf{0.83}  & \textbf{0.33}  & \textbf{1.19} \\
                \bottomrule
            \end{tabular}
            \begin{tablenotes}
                \small
                \item The best score in each category is in bold. Model BO-QGAN is the optimized model from this study. The recreated and proposed models record the average over 30 trials.
            \end{tablenotes}
        \end{threeparttable}
\end{table*}

\subsection{Characteristics of Optimized Architecture}
\begin{figure}[ht]
    \centering
    \includegraphics[width=\columnwidth]{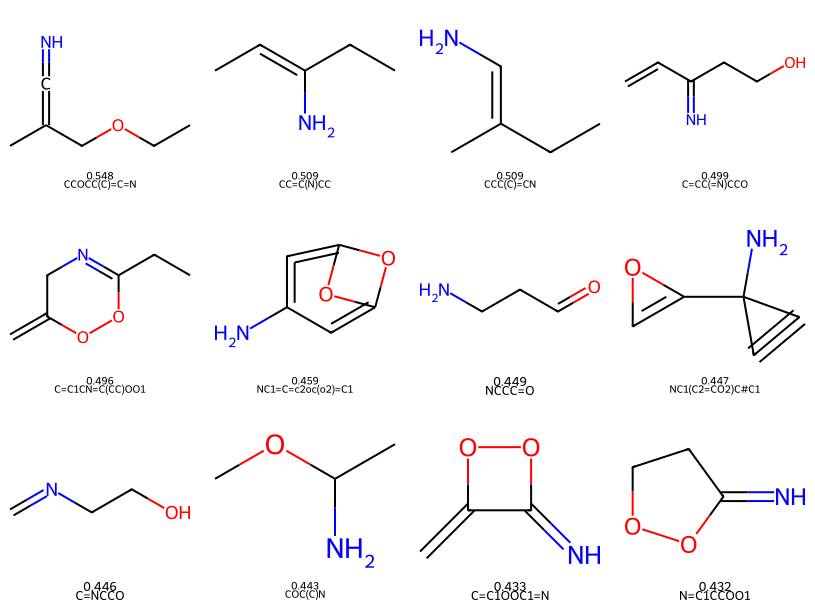}
    \caption[A sample of 12 generated molecules.]{A sample of 12 molecules generated during the inference stage and drawn with RDKit. The value shown is the $QED$ score of the molecule.}
    \label{fig:mols}
\end{figure}

The optimization revealed complex, non-linear relationships between model capacity and performance. The best architecture, Bayesian-Optimized Quantum GAN (BO-QGAN), uses 3 quantum circuit layers (7 qubits each) feeding into 2 classical layers (227 neurons each) (Table \ref{table:architectures}). Its superior $DCS$ is mainly due to improved lipophilicity ($logP$) and synthetic accessibility ($SA$) compared to baselines (Table \ref{table:drug_properties}). A sample of molecules generated by a model and their corresponding $QED$ scores is shown in Figure \ref{fig:mols}.

\begin{table}[htb]
    \centering
    \caption{Model architecture with best performance.}
        \begin{tabular}{ccccc}
            \hline
            \textbf{Qubits} & \textbf{Q Circuits} & \begin{tabular}[x]{@{}c@{}}\textbf{Classical}\\\textbf{Width}\end{tabular} & \begin{tabular}[x]{@{}c@{}}\textbf{Classical}\\\textbf{Layers}\end{tabular} \\
            \hline
            7  & 3  & 227  & 2 \\
            \hline
        \end{tabular}
    \label{table:architectures}
\end{table}

\subsection{Influence of Quantum Architecture} \label{sec:quant-results}
\begin{figure}[htp]
    \centering
        \begin{subfigure}{\columnwidth}
            \centering
            \includegraphics[width=\columnwidth]{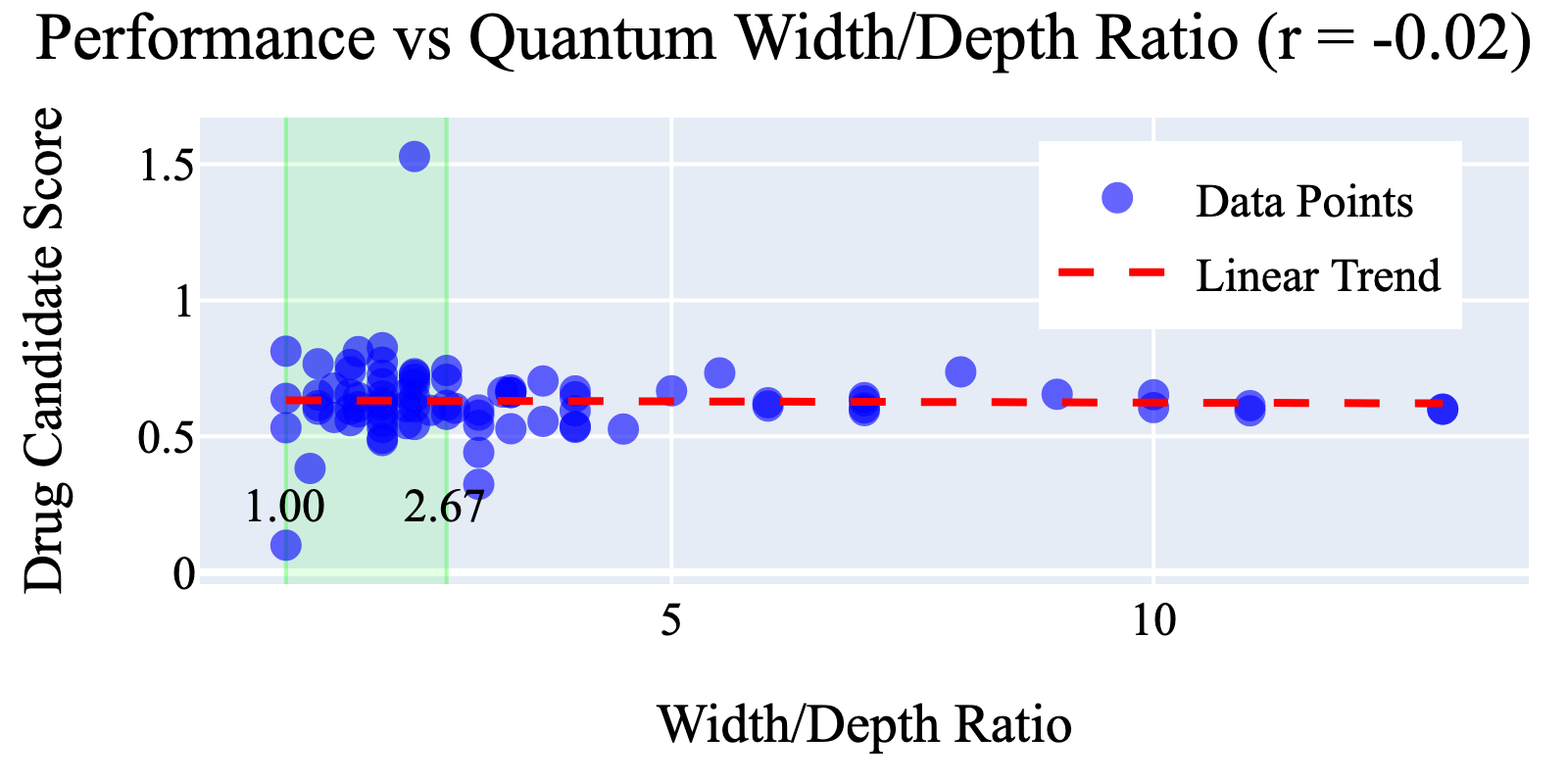}
            \caption[Width-to-depth ratio of quantum circuits vs. $DCS$.]{Comparing the width-to-depth ratio of the quantum circuits to the $DCS$ observed.}
            \label{fig:q_ratio}
        \end{subfigure}

    \vspace{1em}

        \begin{subfigure}{\columnwidth}
            \centering
            \includegraphics[width=\columnwidth]{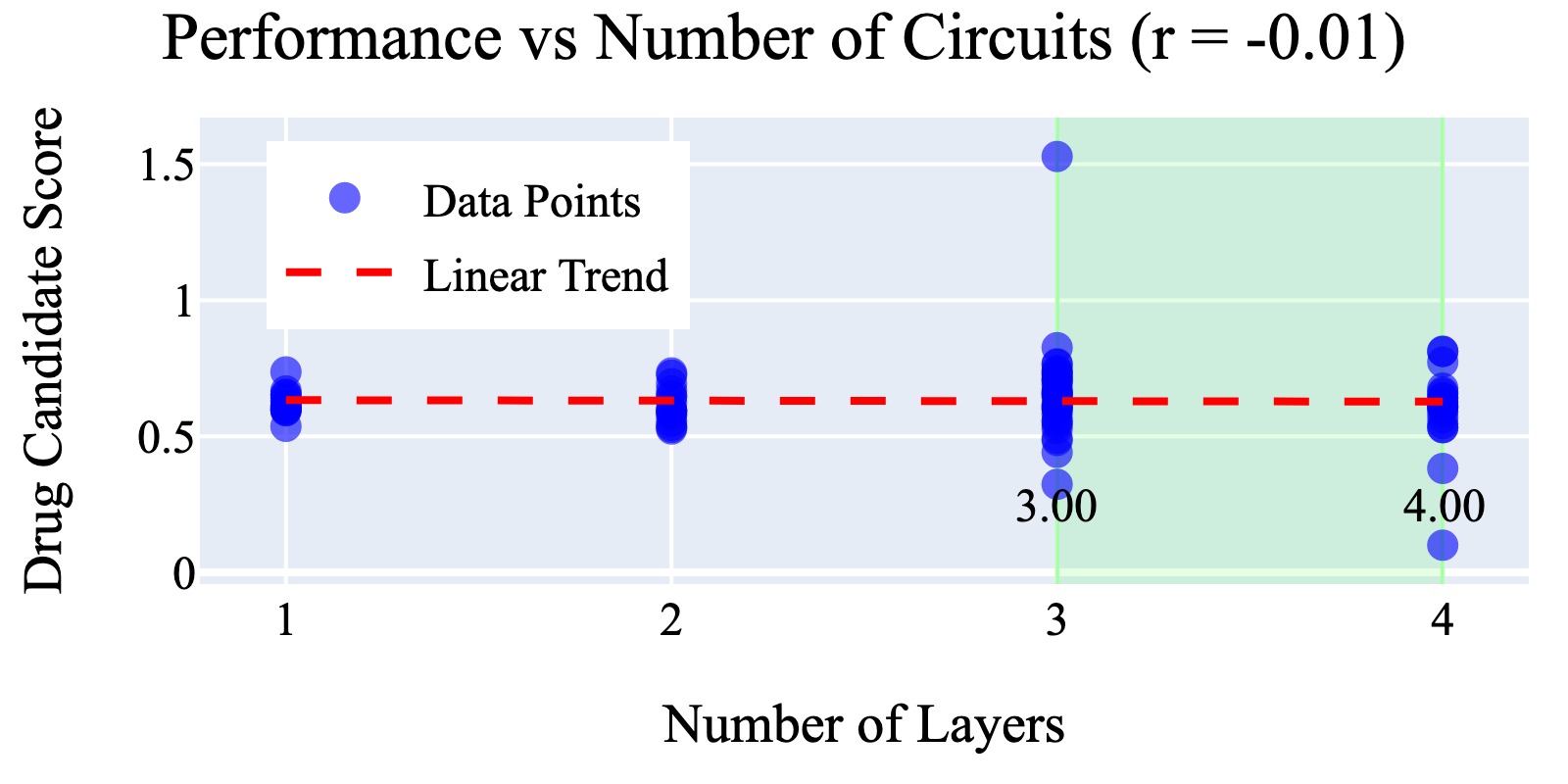}
            \caption[Number of quantum circuits vs. performance.]{Comparing the number of quantum circuits (layers) to the performance of the model.}
            \label{fig:perf_layers_quant}
        \end{subfigure}

    \vspace{1em}

        \begin{subfigure}{\columnwidth}
            \centering
            \includegraphics[width=\columnwidth]{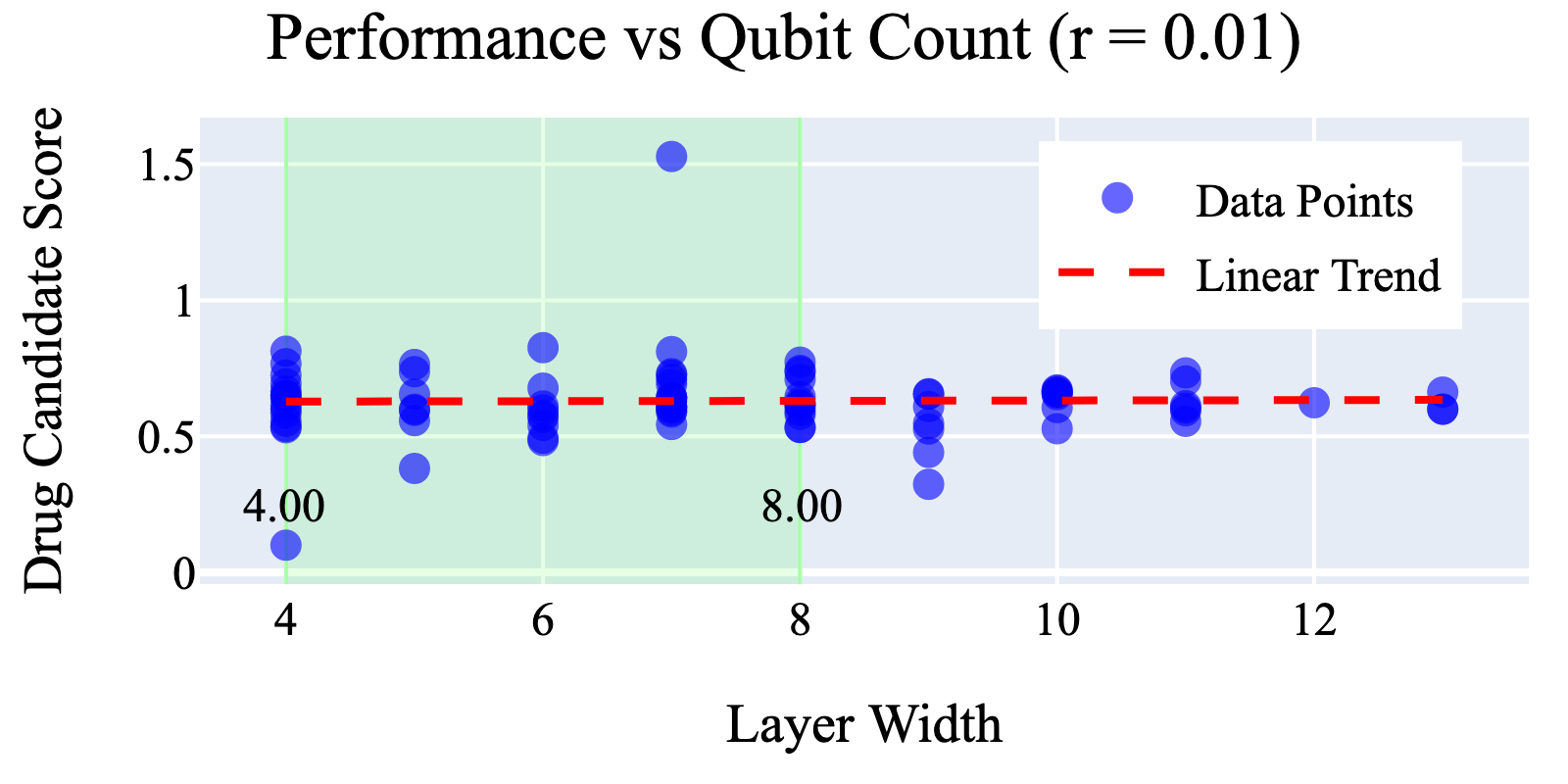}
            \caption[Number of qubits vs. $DCS$.]{Comparing the number of qubits used (width) to the $DCS$. The red line and r score show the linear correlation between variables. The green region contains all the configurations within the top 10\% of $DCS$.}
            \label{fig:perf_width_quant}
        \end{subfigure}

    \caption{Analysis of quantum circuit architectures.}
    \label{fig:q_width_depth}
\end{figure}

The quantum component design was critical. Performance peaked within a middle range of total quantum parameters, suggesting a trade-off between expressivity and computational cost/overfitting. Analysis of width vs. depth (Figure \ref{fig:q_width_depth}) revealed that the best models consistently used multiple (3-4) shallow (4-8 qubits) quantum circuits layered sequentially, corresponding to width-to-depth ratios between $\approx 1.0$ and $2.7$. This indicates that layering more smaller circuits is more effective than using fewer, wider circuits for this task.

\subsection{Influence of Classical Architecture} \label{sec:classical-results}
\begin{figure}[htp]
    \centering
        
        \begin{subfigure}{\columnwidth}
            \centering
            \includegraphics[width=\columnwidth]{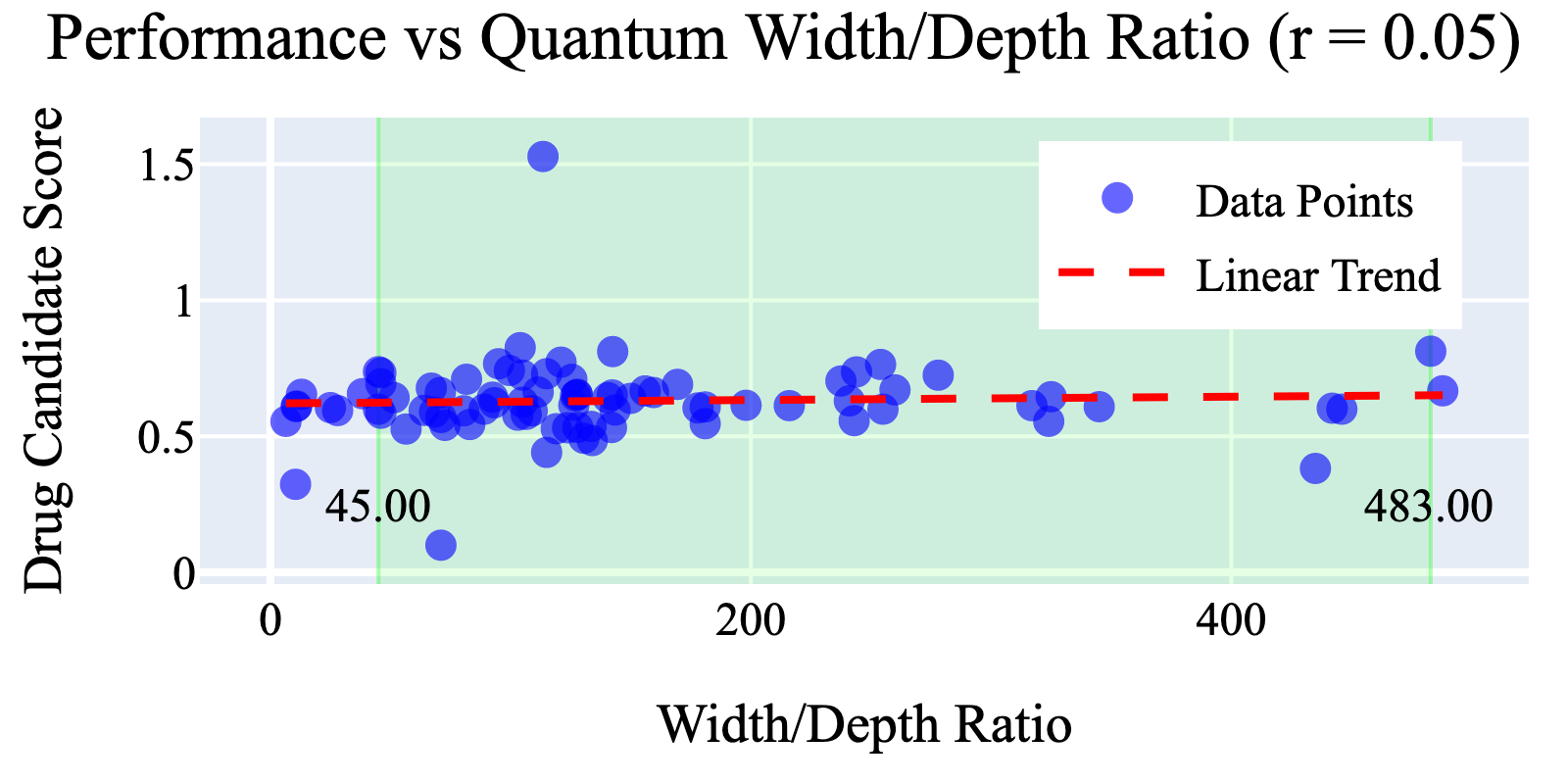}
            \caption[Width-to-depth ratio of classical neural networks vs. $DCS$.]{Comparing the width-to-depth ratio of the classical neural network to the $DCS$ observed.}
            \label{fig:c_ratio}
        \end{subfigure}

        \vspace{1em}

        \begin{subfigure}{\columnwidth}
            \centering
            \includegraphics[width=\columnwidth]{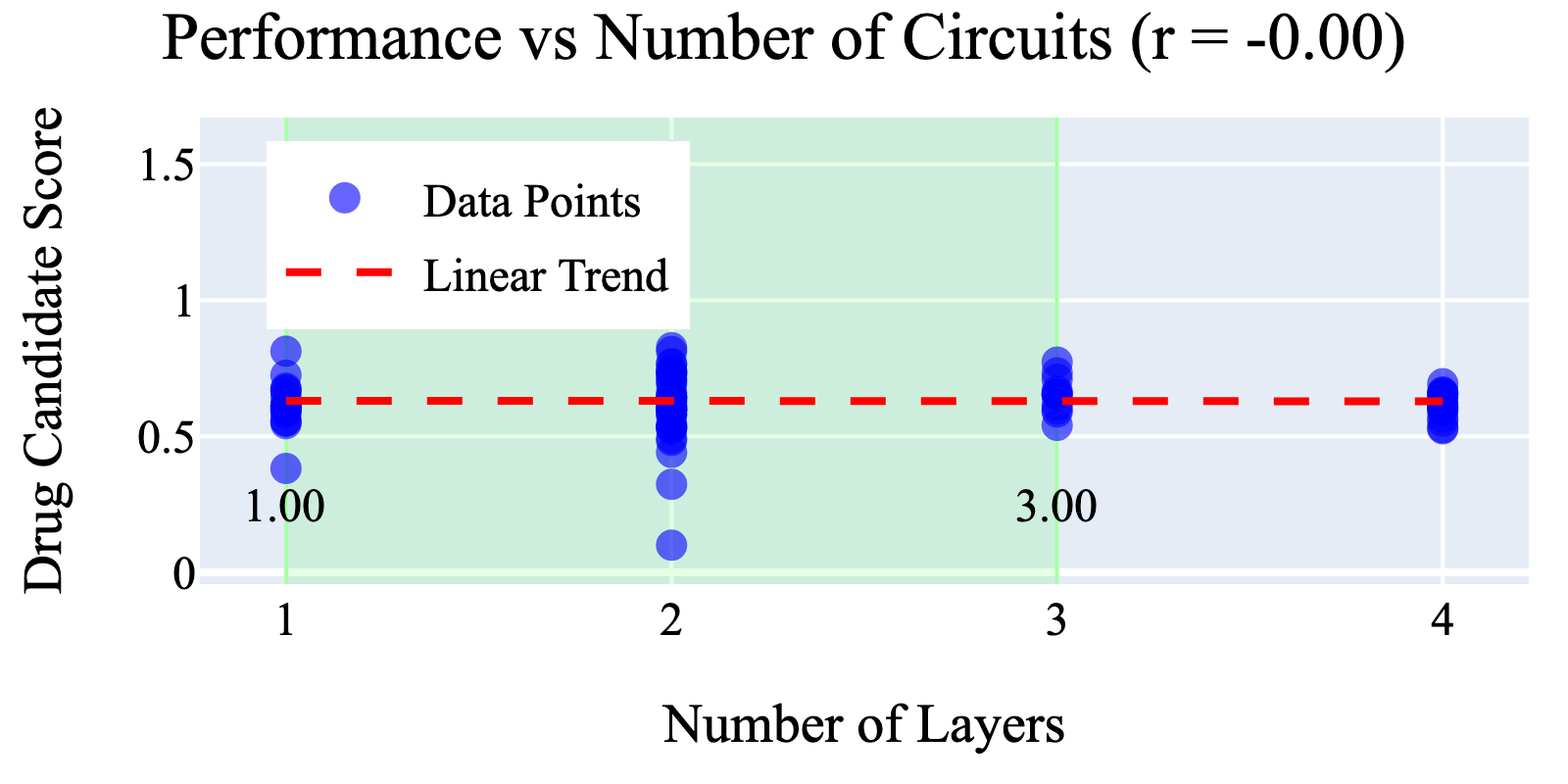}
            \caption[Number of classical layers vs. performance.]{Comparing the number of classical neural network layers to the performance of the model.}
            \label{fig:perf_layers_class}
        \end{subfigure}

        \vspace{1em}

        \begin{subfigure}{\columnwidth}
            \centering
            \includegraphics[width=\columnwidth]{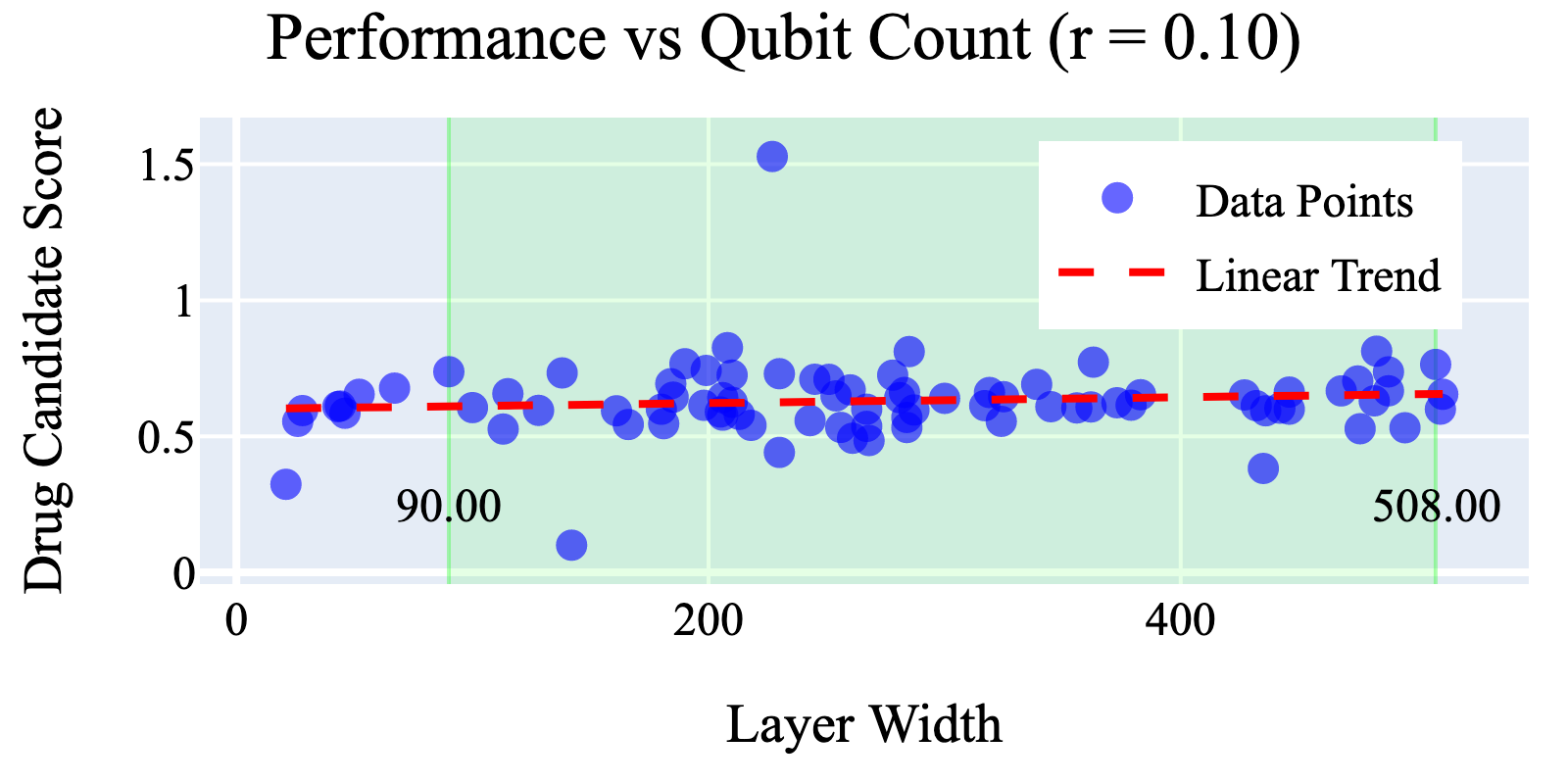}
            \caption[Number of neurons vs. $DCS$]{Comparing the number of nodes per network layer to the $DCS$. The red line and r score show the linear correlation between variables. The green region contains all the configurations with the top 10\% of $DCS$.}
            \label{fig:perf_width_class}
        \end{subfigure}
    \caption{Analysis of classical neural network architectures.}
    \label{fig:c_width_depth}
\end{figure}

In contrast, the classical architecture showed less sensitivity beyond a minimum capacity threshold (approx. $>$90 neurons per layer). Increasing classical parameters or varying the width-to-depth ratio showed little correlation with improved $DCS$ once this threshold was met (Figure \ref{fig:c_width_depth}). This suggests that the classical network primarily needs sufficient capacity to process the quantum output, with diminishing returns from further size increases.

\subsection{Interplay between Quantum and Classical Architecture} \label{sec:interplay}
\begin{figure}[h]
    \centering
    \includegraphics[width=\columnwidth]{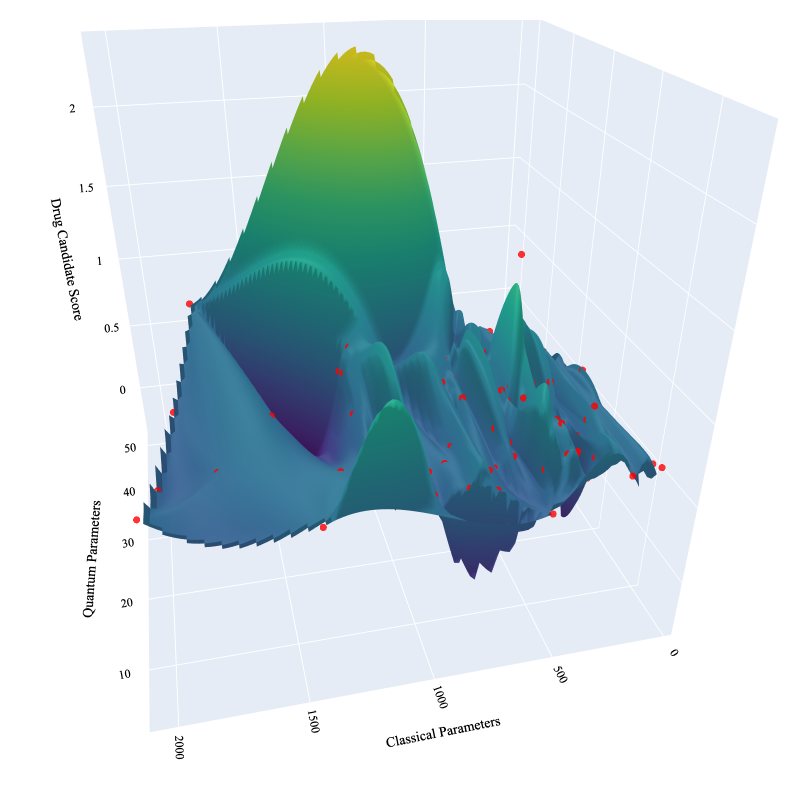}
    \caption[$DCS$ vs. size of quantum circuits and classical neural networks.]{A performance landscape showing how $DCS$ varies with the size of quantum circuits and classical neural networks. Red dots indicate individual trials, and the surface between them is interpolated to visualize trends.}
    \label{fig:qc_interplay}
\end{figure}

Analysis revealed no strong interplay between optimal quantum and classical network sizes. Figure \ref{fig:qc_interplay} visualizes the $DCS$ achieved across the search space (defined by quantum and classical parameter counts). The landscape appears noisy, without clear ridges which would indicate that certain quantum sizes work best only with specific classical sizes. The correlation between the ratio of quantum-to-classical parameters and $DCS$ was weak ($r = -0.20$). Optimal performance seems to depend on finding suitable configurations within each component independently, rather than a specific balance between them.

\subsection{Computational Efficiency}
\begin{figure}[h]
    \centering
    \includegraphics[width=\columnwidth]{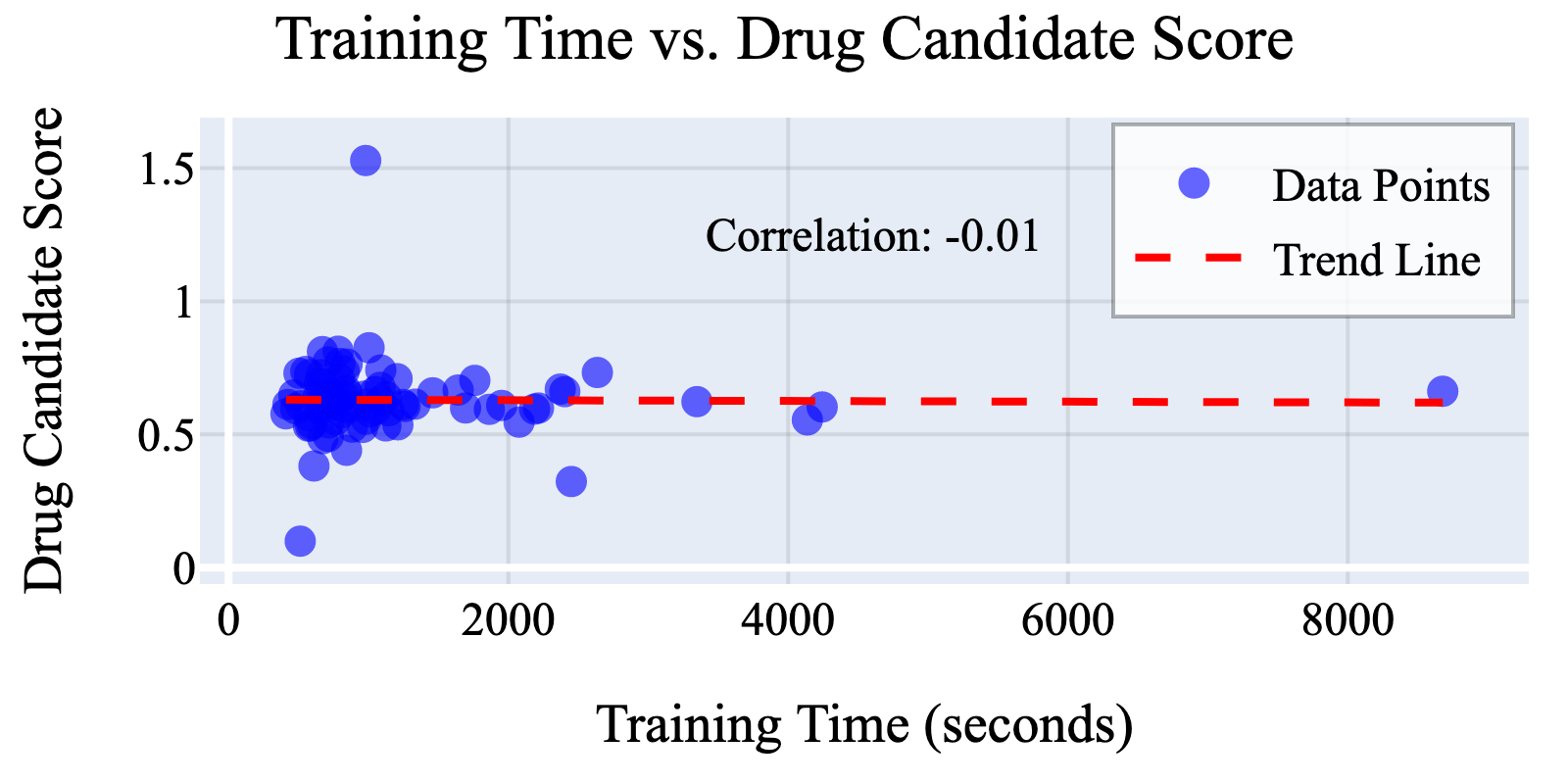}
    \caption{Comparison between training time for a model and $DCS$ achieved.}
    \label{fig:time_vs_dcs}
\end{figure}

The optimization process successfully balanced performance and computational cost. Figure \ref{fig:time_vs_dcs} plots the final $DCS$ achieved against the total training time for each of the 100 trials. No significant correlation was found between longer training times and better final $DCS$ ($r = -0.01$). The best model, BO-QGAN, trained efficiently ($<$17 minutes on test hardware, using 2140 training iterations). Training time with the simulated backend was primarily driven by quantum circuit width (qubits), consistent with the exponential scaling of simulation cost with \(N\). The finding that shallower, narrower circuits perform best aligns well with computational tractability.

\subsection{Validation on Real Quantum Hardware}\label{sec:validation}
To bridge the gap between idealized simulations and practical application on current hardware, we conducted a preliminary validation using a physical quantum processing unit (QPU). Acknowledging the significant constraints associated with NISQ device access time, queue waits, and resource limitations for academic research, this validation necessarily used a small sample size.

For this test, we selected the model configuration identified during optimization that generated the highest percentage of chemically valid molecules, rather than the absolute highest $DCS$ model (BO-QGAN). This choice aimed to maximize the number of usable molecules generated per hardware run, making the most efficient use of limited quantum resources by minimizing invalid outputs which receive zero scores across all metrics.

We generated a batch of 10 molecular structures using this high-validity configuration executed on IBM's 127-qubit QPU `Sherbrooke' via the cloud platform. For direct comparison, a batch of 10 molecules was generated using the same model configuration executed on a classical quantum simulator (PennyLane's `default.qubit`). The resulting molecules from both backends were evaluated using the standard druglikeness metrics ($QED$, $logP$, $SA$, $DCS$).

Table \ref{table:hardware} presents the comparison. While the metrics differ slightly between the QPU and simulator runs, the values generally fall within comparable ranges. The small sample size ($n=10$) and the inherent stochasticity of both GAN generation and NISQ hardware execution prevent definitive conclusions about performance parity or the precise impact of hardware noise (e.g., decoherence, gate errors) on generation quality. However, these initial results are encouraging, tentatively suggesting that the architectural principles derived from simulation may translate reasonably well to physical quantum hardware.

\begin{table}[htb]
    \centering
    \caption{Performance Comparison on Real vs. Simulated Quantum Backends ($n = 10$).}
        \label{table:hardware}
        \begin{tabular}{lcccc}
            \toprule
            \textbf{Backend} & \textbf{$QED$} & \textbf{$logP$} & \textbf{$SA$} & \textbf{$DCS$} \\
            \midrule
            IBM QPU (`Sherbrooke') & 0.50 & 0.76 & 0.24 & 0.89 \\
            Quantum Simulator & 0.43 & 0.73 & 0.37 & 1.17 \\
            \bottomrule
        \end{tabular}
\end{table}

\section{Discussion}

\subsection{Interpretation}
This study demonstrates that systematic, multi-objective Bayesian optimization of the quantum-classical interface significantly enhances hybrid generative model performance and efficiency. Our optimized model, BO-QGAN, substantially outperformed hybrid and classical baselines in generating druglike molecules ($DCS$) while using over 60\% fewer parameters than classical MolGAN. This validates MOTPE for tuning complex hybrid systems and yields empirically-grounded design principles. Key findings include the relative independence of optimal quantum and classical component sizes, the critical role of quantum architecture (favoring multiple shallow circuits), and the lower sensitivity of classical architecture beyond a capacity threshold.

\subsection{Design Principles for Quantum Circuits}
Based on the optimization results, particularly the analysis of the influence of quantum architecture (Section \ref{sec:quant-results}), we can propose practical design guidelines for hybrid generative models in this domain. For the QM9 dataset, characterized by small molecules, the best-performing models consistently favored multiple (3-4), sequentially layered, shallow (4-8 qubits) quantum circuits over fewer wider or deeper ones. This suggests a core principle: when designing the quantum component, prioritize layering relatively shallow circuits sequentially.

The potential reasons for the effectiveness of this layered, shallow approach are multifaceted. It may relate to improved trainability on NISQ devices; shallower circuits are generally less susceptible to noise accumulation and may mitigate the risk of encountering barren plateaus in the optimization landscape \cite{mcclean_barren_2018}. Furthermore, layering circuits could allow the model to learn hierarchical representations, with earlier layers capturing basic features and later layers combining them into more complex patterns, analogous to deep classical networks but within the quantum domain. This approach also aligns well with computational efficiency, especially for simulation, where cost scales exponentially with qubit count (width) but only polynomially with gate depth (layers). Therefore, when allocating limited quantum resources, prioritizing adding circuit layers over adding qubits appears to be a more cost-effective strategy for improving performance on this type of task.

Regarding the classical component, the principle appears simpler: ensure sufficient capacity (roughly $>$90 neurons per layer in our experiments) to process the information from the quantum circuits, after which further increases yield diminishing returns (Section \ref{sec:classical-results}). These empirically derived principles offer valuable, quantitative starting points for designing effective and resource-aware quantum-classical hybrid models in the NISQ era.

\subsection{Limitations and Mitigation}
Key limitations include:
\begin{itemize}
    \item \textbf{Simulation vs. Real Hardware:} Primary results rely on idealized simulations lacking NISQ noise and topology constraints. While preliminary hardware validation (Section \ref{sec:validation}) was encouraging, extensive testing on physical QPUs is needed to confirm real-world viability.
    \item \textbf{QM9 Dataset Scope:} QM9 is a standard benchmark in this field, but it contains small, simple molecules. Generalizability to larger, more diverse drug candidates (e.g., from ZINC or ChEMBL) requires further investigation.
    \item \textbf{MolGAN Baseline Framework:} Performance gains are relative to the MolGAN framework. A simple model was chosen to help isolate the effect of the architecture optimization. Applying the optimization methodology and derived principles to state-of-the-art generative models (e.g. diffusion) is a logical next step.
    \item \textbf{$DCS$ Metric and Optimization Scope:} The $DCS$ metric is one way to quantify quality; results might differ with other objectives. The optimization experiment focused on network width and depth; exploring different gates, embeddings, or measurements could yield further gains.
\end{itemize}

\subsection{Future Research Directions}
Future work should prioritize extensive validation on diverse NISQ hardware platforms with larger sample sizes and error mitigation. Applying the optimization methodology to larger datasets (ZINC, ChEMBL) and more advanced generative frameworks (diffusion, flow-matching) is crucial for assessing scalability and generalizability. Further investigation into optimizing finer architectural details (gates, embedding, measurement) and adapting the framework for target-specific objectives (binding affinity) are also promising directions.

\subsection{Conclusion}
We presented the first systematic multi-objective optimization of the quantum-classical interface for hybrid molecular generative models. Our approach identified architectures yielding significant performance improvements with reduced parameter counts. The resulting BO-QGAN model, validated optimization strategy, and derived design principles (preferring layered, shallow quantum circuits) provide a practical foundation for rationally designing more effective hybrid models. This work helps bridge the gap between quantum potential and drug discovery challenges, potentially accelerating the integration of NISQ computing into pharmaceutical research.

\section{Data and Code Availability}
Data and code are available on \href{https://github.com/amerorchis/Hybrid-Quantum-GAN}{Github (https://github.com/amerorchis/Hybrid-Quantum-GAN)}.

\bibliography{references/references}
\bibliographystyle{icml2025}




\end{document}